\documentclass[runningheads]{llncs}
\usepackage{graphicx}
\usepackage{amsmath}
\usepackage{amsfonts}
\usepackage{tikz}
\usepackage{verbatim}
\usepackage[utf8]{inputenc}
\usepackage[T1]{fontenc}
\usepackage{pgfplots}
\usepackage{hyperref}
\usepackage{booktabs}
\usepackage{placeins}
\DeclareUnicodeCharacter{2212}{−}

\usepgfplotslibrary{groupplots,dateplot}
\pgfplotsset{compat=newest}

\usetikzlibrary{shapes}
\usetikzlibrary{shapes.geometric}
\usetikzlibrary{shapes.arrows}
\usetikzlibrary{arrows.meta}
\usetikzlibrary{patterns}
\usetikzlibrary{positioning}
\usetikzlibrary{fit}
\usetikzlibrary{decorations.pathreplacing}
\usetikzlibrary{calligraphy}
\usetikzlibrary{calc}
\usetikzlibrary{backgrounds}

\DeclareMathOperator{\expectation}{\mathbb{E}}
\DeclareMathOperator{\var}{Var}
\DeclareMathOperator{\cov}{Cov}
\DeclareMathOperator{\mutual}{I}
\DeclareMathOperator{\entropy}{H}
\DeclareMathOperator{\categorical}{Cat}

\DeclareMathOperator{\normal}{\mathcal{N}}

\newcommand*\centermathcell[1]{\omit\hfil$\displaystyle#1$\hfil\ignorespaces}

\definecolor{crimson2143940}{RGB}{214,39,40}
\definecolor{darkgray176}{RGB}{176,176,176}
\definecolor{darkorange25512714}{RGB}{255,127,14}
\definecolor{forestgreen4416044}{RGB}{44,160,44}
\definecolor{steelblue31119180}{RGB}{31,119,180}
\definecolor{mediumpurple148103189}{RGB}{148,103,189}

\tikzset{%
    >={Latex[width=1.5mm,length=1.5mm]},
    every picture/.style={/utils/exec={\sffamily}},
    data/.style = {text centered, fill=white!30, align=center}, 
    txt/.style = {text centered, align=center},
    commonbase/.style = {draw=black, distance=2cm and 6cm, align=center, text centered}, 
    base/.style = {rectangle, commonbase, minimum width=2.5cm, minimum height=2cm},
    modelinner/.style = { fill=steelblue31119180, draw=black,text=white},
    model/.style   = {base, modelinner},
    img/.style   = {data},
}

\hypersetup{
    pdfauthor={Leonhard F. Feiner et al.},
    pdftitle={Propagation and Attribution of Uncertainty in Medical Imaging Pipelines},
    pdfkeywords={uncertainty propagation, uncertainty quantification, Monte Carlo sampling}
}

\begin{document}
\title{Propagation and Attribution of Uncertainty in Medical Imaging Pipelines}
\authorrunning{L. F. Feiner et al.}

\author{
    Leonhard F. Feiner\inst{1,2} 
    \and Martin J. Menten\inst{2,3}
    \and Kerstin Hammernik\inst{2}
    \and \\ Paul Hager\inst{2}
    \and Wenqi Huang\inst{2}
    \and Daniel Rueckert\inst{2,3}
    \and Rickmer F. Braren\inst{1,4}
    \and Georgios Kaissis\inst{1,2,5}
}
\institute{
    Institute of Diagnostic and Interventional Radiology, Klinikum rechts der Isar, Technical University of Munich
    \and AI in Medicine and Healthcare, Klinikum rechts der Isar, 
    Technical University of Munich
    \and BioMedIA,
    Imperial College London
    \and German Cancer Consortium DKTK, Partner Site Munich
    \and Institute for Machine Learning in Biomedical Imaging, Helmholtz-Zentrum Munich
    \email{\{leo.feiner\}@tum.de}
}
\maketitle
\begin{abstract}
Uncertainty estimation, which provides a means of building explainable neural networks for medical imaging applications, have mostly been studied for single deep learning models that focus on a specific task. In this paper, we propose a method to propagate uncertainty through cascades of deep learning models in medical imaging pipelines. This allows us to aggregate the uncertainty in later stages of the pipeline and to obtain a joint uncertainty measure for the predictions of later models. Additionally, we can separately report contributions of the aleatoric, data-based, uncertainty of every component in the pipeline. We demonstrate the utility of our method on a realistic imaging pipeline that reconstructs undersampled brain and knee magnetic resonance (MR) images and subsequently predicts quantitative information from the images, such as the brain volume, or knee side or patient's sex. We quantitatively show that the propagated uncertainty is correlated with input uncertainty and compare the proportions of contributions of pipeline stages to the joint uncertainty measure.
\keywords{uncertainty propagation\and uncertainty quantification \and Monte Carlo sampling}
\end{abstract}
\section{Introduction}
Deep learning has become the state-of-the-art tool for the reconstruction, segmentation and interpretation of medical images. When applied to clinical practice, multiple deep learning models are commonly combined in a cascade of tasks across the imaging pipeline. Hereby, the output of an upstream model is subsequently used as input of a downstream model. For example, deep learning may be used to first reconstruct magnetic resonance (MR) images from raw k-space data before the reconstructed images are interpreted by another algorithm for signs of disease.
At the same time, the application of deep learning models in clinical practice requires an estimate of their uncertainty. Ideally, the algorithm informs its user about unsure predictions in order to prevent harmful medical decisions based on incorrect predictions~\cite{tschandl2020human}. Many solutions to estimate the uncertainty of a single deep learning model have been introduced~\cite{blundell2015weight,Gal2016,Kendall2017a}.

Integrating uncertainty estimation with imaging pipelines consisting of cascading deep learning models comes with additional challenges. The uncertainty of upstream models directly affects the output of downstream models. The \emph{propagation of uncertainty} through the cascade has to be explicitly modeled in order to obtain a \emph{joint} uncertainty measure for the entire pipeline. This also allows for \emph{attribution} of the total uncertainty to the pipeline's individual components. To address these unmet needs, we make the following contributions in this work:

\begin{itemize}
    \item We propose a novel method to propagate uncertainty through a pipeline with multiple deep learning components using Monte Carlo sampling.
    \item The proposed strategy
    allows the calculation of a joint uncertainty measure for the whole pipeline, and the attribution of the contributions to the pipeline's individual models for both classification and regression tasks.
    \item The utility of 
    the proposed strategy is demonstrated on realistic medical image processing pipelines, in which the upstream models reconstruct magnetic resonance (MR) images from undersampled k-space data with varying, controllable amounts of aleatoric uncertainty. Different downstream models predict the brain volume, the knee side or the patient's sex. The code is available at \hyperlink{https://github.com/LeonhardFeiner/uncertainty\_propagation}{github.com/LeonhardFeiner/uncertainty\_propagation}.
\end{itemize}
\section{Related Work}
In general, two sources of uncertainty can be distinguished: \emph{epistemic} (or model) uncertainty and \emph{aleatoric} uncertainty, i.e. noise and missing information in the data~\cite{Kendall2017a}. Recently, Bayesian methods have been developed to estimate epistemic uncertainty in machine learning models, such as Dropout during inference~\cite{Gal2016}, learning weight distributions using backpropagation~\cite{blundell2015weight}, and model ensembling~\cite{lakshminarayanan2017simple}. To estimate the aleatoric uncertainty, previous works have suggested modeling the deterministic network output and intermediate activation functions by distributions~\cite{Gasta}, to perform test-time augmentation~\cite{wang2019aleatoric} or estimating the mean and variance of the target distribution~\cite{Nix1994}. Shaw et al. separate aleatoric uncertainty sources by removing components during training~\cite{shaw2020heteroscedastic}. 

Uncertainty estimation has been applied to various tasks in medical image processing. In MR image reconstruction, pixel-wise epistemic uncertainty was estimated by drawing model parameters from a learned multivariate Gaussian distribution~\cite{Narnhofer2022} or applying posterior sampling using Langevin Dynamics with a deep generative prior~\cite{jalal2021robust}. Another approach used Monte Carlo dropout and a heteroscedastic loss to model aleatoric and epistemic uncertainty~\cite{Schlemper2018}. 
Many works have evaluated uncertainty in medical image classification~\cite{ghoshal2021estimating,ju2022improving} and regression~\cite{feng2020bayesian,laves2020well}. Uncertainty estimation has also been integrated with models for MR image super-resolution~\cite{Tanno2017} or medical image registration~\cite{Dalca2019}. However, all these works are limited to the uncertainty estimation of a single model and do not consider a cascade of models across a typical imaging pipeline.

Techniques for uncertainty propagation include Monte Carlo sampling~\cite{HussenAbdelaziz2015a,Ji,wang2019aleatoric}, unscented transforms~\cite{Astudillo2011,Novoa2018}, linearizing the non-linearities of the network 
partially~\cite{Bibi} or fully~\cite{Postels2019,Titensky2018},
as well as performing assumed density filtering~\cite{Gasta,Ghosh2016}. They estimate uncertainty by assuming constant image noise as input uncertainty~\cite{Bibi,Gasta,Ji,Titensky2018}, generating the uncertainty within the model layers~\cite{Gasta,Ghosh2016,Postels2019,Titensky2018}, interpreting augmentations as distribution~\cite{wang2019aleatoric}, or using the output of classical language recognition models~\cite{HussenAbdelaziz2015a,Astudillo2011,Novoa2018}. None of these works use a pipeline of deep learning models or combine the predicted uncertainty of multiple models into a joint uncertainty metric.

The following works have specifically investigated uncertainty estimation in medical imaging pipelines~\cite{Mehta2021,mehta2019propagating,ozdemir2017propagating}. They use a cascade of models that each output uncertainty estimations in addition to their prediction. However, their methods cannot quantify the influence of the uncertainty of upstream tasks on the uncertainty of the downstream task. This is because their methods concatenate uncertainty maps and the prediction as input channels for downstream models. Consequently, it is impossible to attribute the output uncertainty to either the input uncertainty and individual components of the pipeline. Revealing this dependence would require the propagation of probability distributions through the network, which we propose in our work.

\section{Methods}
\begin{figure}[!b]
\centering
\input{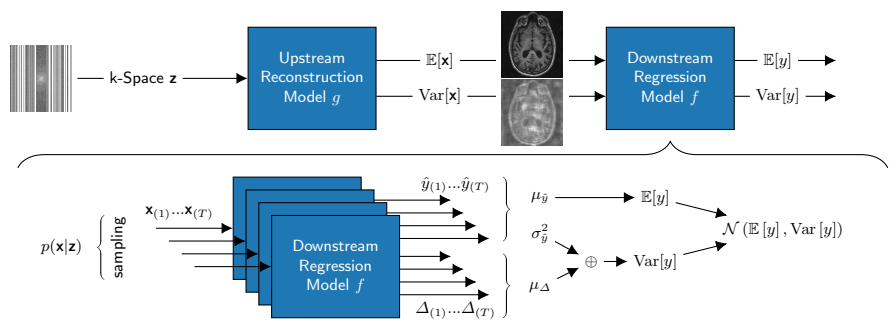}\\
\begin{tikzpicture}[scale=0.68, every node/.style={scale=0.68}]
{
    \node (downstreamm1)   [model]   { Downstream\\ Regression\\ Model $f$};
    \draw[->]  ([yshift=0.55cm]downstreamm1.east) -- ++(1.75, 0);
    \draw[->]  ([yshift=-0.55cm]downstreamm1.east) -- ++(1.75, 0);
    \node (downstreamm2)   [model,  below of=downstreamm1, xshift=0.25cm, yshift=0.75cm]   { Downstream\\ Regression\\ Model $f$};
    \draw[->]  ([yshift=0.55cm]downstreamm2.east) -- ++(1.75, 0);
    \draw[->]  ([yshift=-0.55cm]downstreamm2.east) -- ++(1.75, 0);
    \node (downstreamm3)   [model,  below of=downstreamm2, xshift=0.25cm, yshift=0.75cm]   { Downstream\\ Regression\\ Model $f$};
    \draw[->]  ([yshift=0.55cm]downstreamm3.east) -- ++(1.75, 0);
    \draw[->]  ([yshift=-0.55cm]downstreamm3.east) -- ++(1.75, 0);
    \node (downstreamm4)   [model,  below of=downstreamm3, xshift=0.25cm, yshift=0.75cm]   { Downstream\\ Regression\\ Model $f$};
    \draw[->]  ([yshift=0.55cm]downstreamm4.east) -- ++(1.75, 0);
    \draw[->]  ([yshift=-0.55cm]downstreamm4.east) -- ++(1.75, 0);
    \draw[->] ([xshift=-1.5cm]downstreamm1.west) -- (downstreamm1);
    \draw[->] ([xshift=-1.5cm]downstreamm2.west) -- (downstreamm2);
    \draw[->] ([xshift=-1.5cm]downstreamm3.west) -- (downstreamm3);
    \draw[->] ([xshift=-1.5cm]downstreamm4.west) -- (downstreamm4);

    \draw [decorate,decoration = {calligraphic brace,mirror,raise=10pt}] ([xshift=2.25cm,yshift=-0.25cm]downstreamm1.east) -- ++(0, 1.25);
    \draw [decorate,decoration = {calligraphic brace,raise=10pt}] ([xshift=2.25cm,yshift=-0.5cm]downstreamm1.east) -- ++(0, -1.25);
   
    \draw [decorate,decoration = {calligraphic brace,raise=10pt, mirror}] ([xshift=-2cm,yshift=0.25cm]downstreamm1.west) -- ++(0, -1.25);
    \node (input_distributution) [data, left of=downstreamm1, xshift=-3.6cm, yshift=-0.375cm] {$p(\textbf{x}|\textbf{z})$};
    \node (sampling) [data, left of=downstreamm1, xshift=-2.5cm, yshift=-0.375cm, rotate=90] {sampling};

    \node (input_sample) [txt, left of=downstreamm1, xshift=-1.3cm, yshift=0.3cm] {$\textbf{x}_{(1)} ... \textbf{x}_{(T)}$};
    \node (input_sample) [txt, right of=downstreamm1, xshift=2.1cm, yshift=0.85cm] {$ \hat{y}_{(1)} ... \hat{y}_{(T)}$};
    \node (input_sample) [txt, right of=downstreamm4, xshift=1.35cm, yshift=-0.85cm] {$\Delta_{(1)} ... \Delta_{(T)}$};

    \node (output_mean) [data, right of=downstreamm1, xshift=3.75cm, yshift=0.6cm] {$\mu_{\hat{y}}$};
    \node (output_sigma) [data, right of=downstreamm1, xshift=3.75cm, yshift=-0.1cm] {$\sigma^2_{\hat{y}}$};
    \node (output_delta) [data, right of=downstreamm4, xshift=3cm, yshift=-0.4cm] {$\mu_{\Delta}$};

    \node (output_expectation) [data, right of=output_mean, xshift=1.25cm] {$\expectation[y]$};
    
    \node (output_addition) [data, right of=output_sigma, xshift=0cm, yshift=-0.55cm] {$\oplus$}; 
    
    \node (output_var) [data, right of=output_sigma, xshift=1.25cm, yshift=-0.55cm] {$\var[y]$};
    \node (output_distribution) [data, right of=output_expectation, xshift=1.5cm, yshift=-0.625cm] {$\normal\left(\expectation\left[y\right], \var\left[y\right]\right)$};

    \draw[->] (output_mean) -- (output_expectation);
    \draw[->] (output_sigma) -- (output_addition);
    \draw[->] (output_delta) -- (output_addition);
    \draw[->] (output_addition) -- (output_var);
    \draw[->] (output_expectation) -- (output_distribution);
    \draw[->] (output_var) -- (output_distribution);
}
\end{tikzpicture}
\caption{An example of an imaging pipeline consisting of an upstream MR image reconstruction model $g$, and a downstream regression model $f$. Both models predict a measure of aleatoric uncertainty. Our method allows for the propagation of the mean and variance outputs of the upstream model through the downstream regression model.} \label{fig:downstream_regression}
\end{figure}

Our novel technique for propagation of aleatoric uncertainty can be applied to a model cascade of arbitrary length. For simplicity, we here limit the explicit presentation of our method to one upstream model $g$ and one downstream model $f$. The latter uses the output of $g$ as input (see Figure \ref{fig:downstream_regression}). In the following, $\textbf{z}$ denotes the input data of the pipeline, $\textbf{x}$ expresses the random variable of possible intermediate outputs of the upstream model, and $y$ is a random variable of possible outputs of the entire pipeline.
Without loss of generality, we assume that $\textbf{z}$ and $\textbf{x}$ are vectors (bold), whereas $y$ is a single variable. Both $\textbf{x}$ and $y$ are associated with a single data point $\textbf{z}$ of the dataset, whereas $p(\textbf{x}|\textbf{z})$ and $p(y|\textbf{z})$ are the distributions of the random variables that are predicted by the pipeline up to a certain model stage. We assume the distribution of $p(y|\textbf{x})$ to be normal in the case of regression and categorical in the case of classification. We estimate the mean and variance of the target normal distributions using the technique by Nix et al.~\cite{Nix1994}, whereas the parameters of categorical distributions are given by softmax outputs. We choose the variance or entropy of $y$, respectively, as an uncertainty measure. In the following, we describe the composition of our pipeline in more detail.
\subsection{Upstream Model} The upstream model $g$ outputs a prediction and its uncertainty in the form of the parameters of a distribution $p(\textbf{x}|\textbf{z})$ from which we can sample. In our case, the upstream model produces images as outputs. We follow the common practice to model image uncertainty as pixel-wise variance~\cite{Kendall2017a}, recognizing that this neglects potential higher order spatial correlations~\cite{monteiro2020stochastic}. Spatial correlations in medical images can be associated with various factors, including similar tissue types across the image, local neighborhoods of voxels, or reconstruction artifacts.
As the model $g$ uses a heteroscedastic loss for training and outputs a tuple of arrays containing the mean image $\expectation\left[x\right]$ and the variance image $\var\left[x\right]$, the image $\textbf{x}$ is distributed as a diagonal multivariate normal distribution over predictions.
\subsection{Downstream Model and Joint Pipeline} Next, the downstream model $f$ processes $\expectation\left[x\right]$ and $\var\left[x\right]$. As part of our contributions, we introduce a method to aggregate the uncertainties of the upstream model $g$ and the downstream model $f$ to a joint uncertainty measure. We 
propagate the uncertainty of the intermediate result, given by the distribution $p(\textbf{x}|\textbf{z})$, and obtain its contribution to the uncertainty of the final prediction $p(y|\textbf{z})$. The joint uncertainty measure can be calculated by marginalizing the distribution of possible predictions $\textbf{x}$ of the upstream model $p(y|\textbf{z}) = \int p(y|\textbf{x}) p(\textbf{x}|\textbf{z}) \mathrm{d}{\textbf{x}}$. We approximate this integral by Monte Carlo sampling. The form of the joint uncertainty is different for regression and classification downstream tasks. We describe both cases below.

\noindent\textbf{Regression Downstream Model:} In the regression case, we assume that the prediction is normally distributed. Its mean is given by the expectation over the output value $\expectation[y]$. Its variance describing the joint aleatoric uncertainty can be denoted as the variance of the output value $\var[y]$. Hence, the plausible predictions $y$ are distributed as $y \sim \normal(\expectation[y],\var[y])$. However, the joint variance of the pipeline is the uncertainty of the upstream model propagated through the downstream model $\var[x_{\text{prop}}]$ and the uncertainty of the downstream model itself $\var[y_{\text{ds}}]$. The uncertainty of the downstream model $\var[y_{\text{ds}}]$ can not easily be computed in closed form, but is in general correlated with the propagated uncertainty of the first model $\var[x_{\text{prop}}]$.

The downstream model $f$ does not output $\expectation[y]$ and $\var[y_{\text{ds}}]$ directly, but rather returns a tuple of outputs $\left(\hat{y}, \Delta \right)$. To perform variance propagation, we obtain $T$ Monte Carlo samples $(\textbf{x}_{(1)}, \dots, \textbf{x}_{(T)})$ from the distribution $p(\textbf{x}|\textbf{z})$.
In Figure \ref{fig:downstream_regression}, the $T$ copies of the downstream model $f$ denote simultaneous forward passes using the same model but with different samples as inputs. We use these Monte Carlo samples to approximate the expectation of the predictive distribution using the empirical mean $\expectation[y]\approx\mu_{\hat{y}}$ and the joint variance $\var[y]$ by the empirical variance $\sigma^2_{\hat{y}}$ and the empirical mean $\mu_{\Delta}$ in the following form:
\begin{alignat*}{2}
\var[y] &={}& \centermathcell{\underbrace{\var[x_{\text{prop}}]}}   &+{}& \centermathcell{\underbrace{\var[y_{\text{ds}}] + 2 \cov[x_{\text{prop}}, y_{\text{ds}}]}} \\
 &={}& \centermathcell{ \underbrace{\frac{1}{T-1}\left(\sum_{t=1}^T\hat{y}^2_{(t)} - \left(\sum_t^T\hat{y}_{(t)}\right)^2\right) }
 }   &+{}& \centermathcell{ \underbrace{ \frac{1}{T}\sum_{t=1}^T\Delta_{(t)} } 
 } \\
 &={}& \centermathcell{ \sigma^2_{\hat{y}} }   &+{}& \centermathcell{ \mu_\Delta } .
\end{alignat*}
\noindent\textbf{Classification Downstream Model:} In the classification case, the prediction of the pipeline is given as a categorical distribution over classes $c$ of a one-hot-encoded vector $y$. The downstream model outputs a vector of class confidences $\hat{y}$. To approximate the pipeline's expectation of the predictive distribution, we calculate the empirical mean of the model output $\expectation[y]\approx\mu_{\hat{y}}$. The resulting categorical distribution specifies a certain class confidence by $p({y}=c|\textbf{x})=\expectation\left[{y}\right]_c$. One measure to express the joint uncertainty of a categorical distribution is the entropy of the prediction. The entropy for this pipeline is calculated as follows:
\begin{align*}
\entropy \left[y|\textbf{z}\right]
&=-\sum^C_{c=1}\expectation\left[p\left({y}=c|\textbf{x}\right)\right] \log\expectation\left[p\left({y}=c|\textbf{x}\right)\right].
\end{align*}
As the joint entropy consists of the combined uncertainty from the upstream and downstream model, we want to separate the part of the uncertainty contributed by the propagation. Hence, the mutual information $\mutual$ of the propagated aleatoric uncertainty and the aleatoric uncertainty of the downstream model is derived from the entropy $\entropy$ as follows: $\mutual \left[y,\textbf{x}\right|\textbf{z}] = \entropy \left[\expectation\left[{y}|\textbf{z}\right]\right]-\expectation\left[\entropy \left[{y}|\textbf{x},\textbf{z}\right]\right]$. Further details can be found in the supplementary material.

\subsection{Loss and Parametrization}
Instead of predicting the variance $\sigma^2$ of the upstream model and the residual uncertainty $\Delta$ of the downstream regression model directly, we reparametrize
$\sigma^2$ as $\sigma^2=e^s$ and $\Delta$ as $\Delta=e^\delta$, where $s$ and $\delta$ are the respective outputs of the models. This ensures that $\sigma^2$ and $\Delta$ are positive. For both training and evaluation of the downstream model, we have to minimize the negative log likelihood of the joint distribution $p(y|\textbf{z})$ for each input data point $\textbf{z}$. Since the input of the downstream task is a probability distribution over images, we empirically chose to take 8 Monte Carlo samples during training and 256 Monte Carlo samples during evaluation to approximate the expectation. While the lower number of samples during training sacrifices accuracy for efficiency, we feel that the increased level of noise is mitigated over the course of repeated forward passes.
\section{Experiments and Results}
We test the utility of our method for three different medical pipelines with up- and downstream tasks. In all cases, the upstream task is to reconstruct MR images from undersampled k-space data. Different undersampling rates represent a varying source of aleatoric uncertainty. The downstream task is either a classification or regression task based on the reconstructed images. We demonstrate how our method propagates the uncertainty stemming from different undersampling factors to the ultimate prediction. Additionally, we show how it facilitates the attribution of the joint uncertainty to the individual models of the pipeline.
\begin{figure}[t]
    \centering
    \input{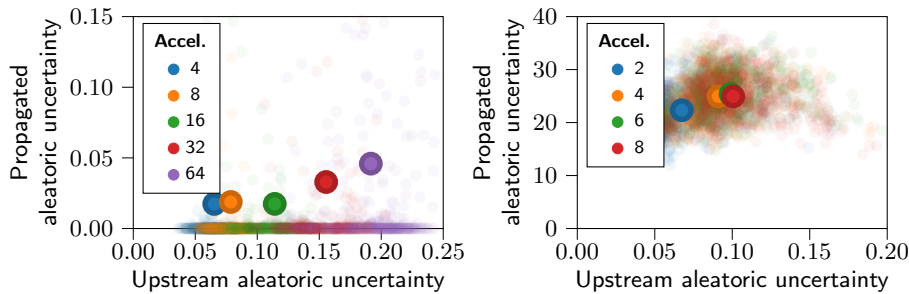}
    \caption{Higher acceleration factors (color) lead to higher reconstruction uncertainty (x-axis) and, through propagation via our method, contribute to a higher uncertainty of the final pipeline output. The propagated portion of the output uncertainty (y-axis) is given as mutual information for knee side classification (left) and as standard deviations of brain volume predictions in ml (right). Large dots are aggregate values.
    }
    \label{fig:scatterboth}
\end{figure}
\noindent\textbf{Reconstruction and Classification of Knee MR Images:} 
\label{section:result_knee}
Our first experiments are based on the fastMRI single-coil knee MR dataset~\cite{Zbontar2018}. The dataset contains reference images and raw k-space data, which we undersample with varying acceleration rates (Accel.) of 4, 8, 16, 32 and 64 by randomly removing columns from the k-space data. A fraction of the most centered columns in k-space is always used during reconstruction (C. Frac.). We use a physics-based reconstruction model based on an unrolled neural network~\cite{Schlemper2017,Schlemper2018}. In addition to the reconstructed 2D MR image, the model also outputs a map of the heteroscedastic aleatoric uncertainty. The upstream model's output is subsequently processed by a downstream model, a modified ResNet-50, that classifies the side of the knee (i.e. left or right knee). Based on the outputs of the downstream model, the pipeline calculates the parameters of a joint categorical distribution over possible predictions containing the uncertainty information. We use the fastMRI validation set containing 199 images as test set and split the original training set containing 973 images patient-wise into two parts to train the upstream and downstream model, respectively. Each model is trained with 80\% of its data split and validated on the remaining 20\%.

\begin{figure}[t]
\centering
\input{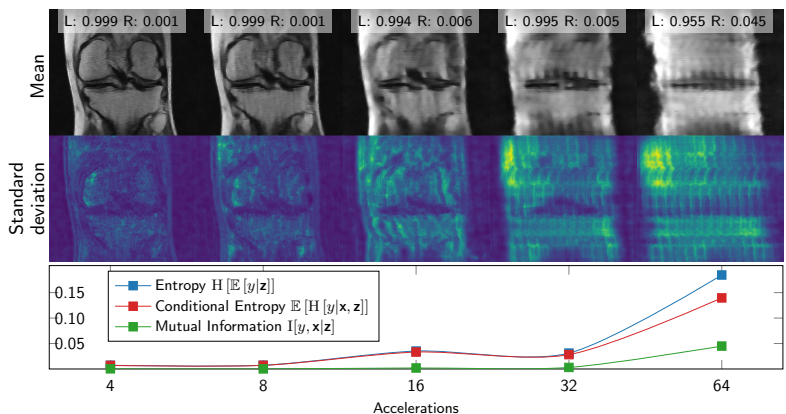}
\caption{Representative example (left knee) of a reconstructed knee MR images from k-space data with varying undersampling factors. Shown are the mean (top) and standard deviation maps (middle) as well as the accompanying uncertainty measures of the downstream network's knee side predictions (bottom).}
\label{knee_overview}
\end{figure}
\begin{table}[t]
\caption{Quantitative measures characterizing the input k-space data, reconstructed images from the upstream model and final output from the downstream model.
}\label{kneemc256}
\centering
\begin{tabular}{c c|c c|c c c c}
\multicolumn{2}{c|}{k-space} & \multicolumn{2}{c|}{Reconstructed image} & \multicolumn{4}{c}{Classification output (knee side)} \\
Accel.& C. Frac.&\hphantom{---}SSIM\hphantom{---}& $\sqrt{\var[\textbf{x}]}$&\hphantom{---}ACC\hphantom{---}& $\mutual \left[y,\textbf{x}\right|\textbf{z}]$ & \hphantom{-}
$\expectation\left[\entropy \left[{y}|\textbf{x},\textbf{z}\right]\right]$
\hphantom{-}
&$\entropy \left[\expectation\left[{y}|\textbf{z}\right]\right]$ \\
\hline
\hphantom{0}4  &0.16 &0.823 & 0.065 & 0.992 & 0.32 $\cdot 10^{-2}$ & 1.63 $\cdot 10^{-2}$ & 1.95 $\cdot 10^{-2}$ \\
\hphantom{0}8  &0.08 &0.757 & 0.079 & 0.997 & 0.32 $\cdot 10^{-2}$ & 1.40 $\cdot 10^{-2}$& 1.71 $\cdot 10^{-2}$ \\
16 &0.04 &0.674 & 0.114 & 0.992 & 0.34 $\cdot 10^{-2}$ & 1.37 $\cdot 10^{-2}$ & 1.71 $\cdot 10^{-2}$ \\
32 &0.02 &0.556 & 0.156 & 0.982 & 0.76 $\cdot 10^{-2}$ & 2.14 $\cdot 10^{-2}$ & 2.90 $\cdot 10^{-2}$ \\
64 &0.01 &0.445 & 0.192 & 0.967 & 1.74 $\cdot 10^{-2}$ & 5.20 $\cdot 10^{-2}$ & 6.93 $\cdot 10^{-2}$ \\
\end{tabular}
\end{table}

We observe that both the uncertainty in the reconstructed images and the propagated uncertainty in the final prediction increase with higher undersampling factors (see Figure \ref{fig:scatterboth} left). Figure \ref{knee_overview} illustrates this effect for a representative, single sample. These observations are also reflected in the quantitative results (see Table \ref{kneemc256}). With increasing undersampling, the uncertainty in the data increases, which is reflected in a higher estimated reconstruction uncertainty and lower structural similarity (SSIM) compared to the ground truth image that is obtained using the entire k-space data. The estimated reconstruction uncertainty is given as the square root of the average over the dataset and pixels ($\sqrt{\var[\textbf{x}]}$). This increased uncertainty is propagated by the downstream classification model. Higher upstream uncertainty yields a higher joint entropy ($\entropy \left[\expectation\left[{y}|\textbf{z}\right]\right]$), as well as a higher mutual information ($\mutual \left[y,\textbf{x}\right|\textbf{z}]$) between the joint entropy and propagated uncertainty. For all undersampling factors, the prediction accuracy (ACC) is very high, which is most likely due to the simple task at hand.



\noindent\textbf{Reconstruction of Brain MR Images:}
For the second set of experiments, we use the Alzheimer’s Disease Neuroimaging Initiative (ADNI) brain MR image dataset~\cite{petersen2010alzheimer}. We calculate the complex-valued k-space data by applying a Fourier transform to the images and add Gaussian noise to the synthetic k-space to mimic the MR imaging process.
\begin{figure}[t]
\centering
\input{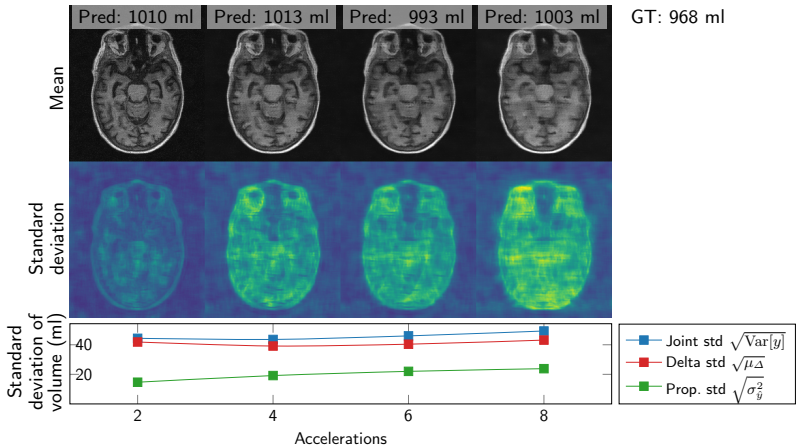}
\caption{Representative example of a reconstructed brain MR images from synthetic k-space data with varying undersampling factors. Shown are the mean (top) and standard deviation maps (middle) as well as the accompanying uncertainty measures in scale of ml of the downstream network's volume predictions (bottom).}
\label{brain_overview}
\end{figure}
\begin{table}[t]
\centering
\caption{Quantitative measures characterizing the input k-space data, reconstructed images from the upstream model and final output from the downstream model.
}
\label{brainmc256}
\begin{tabular}{c c|c c|c c c c c}
\multicolumn{2}{c|}{k-space} & \multicolumn{2}{c|}{Reconstructed image} & \multicolumn{5}{c}{Regression output (brain volume)} \\
Accel.& C. Frac.&\hphantom{---}SSIM\hphantom{---}&  $\sqrt{\var[\textbf{x}]}$  & \hphantom{---}L$_1$\hphantom{---} & \hphantom{---}L$_2$\hphantom{---} & \hphantom{--}$\sqrt{{\sigma^2_{\hat{y}}}}$\hphantom{--} & \hphantom{--}$\sqrt{{\mu_\Delta}}$\hphantom{--} &  $\sqrt{\var[y]}$  \\
\hline
2 &0.16 & 0.714 & 0.0642 & 63.6 & 78.9 & 22.0 & 59.9 & 64.2 \\
4 &0.08 & 0.567 & 0.0880 & 64.3 & 80.9 & 24.6 & 60.7 & 65.9 \\
6 &0.06 & 0.502 & 0.0957 & 64.3 & 81.6 & 25.1 & 62.3 & 67.6 \\
8 &0.04 & 0.450 & 0.0973 & 66.1 & 83.7 & 24.6 & 65.6 & 70.4 \\
\end{tabular}
\end{table}
We again simulate undersampling with acceleration factors of 2, 4, 6, and 8. We perform two different tasks on this dataset: regression of the brain volume and classification of the patient's sex. For the classification task, we use the same pipeline as for the fastMRI dataset, whereas for regression, we change the downstream model's output and loss accordingly. This dataset of 818 MR images is split patient-wise. We use 20\% as a test set and split the remaining 80\% into four subsets to train and validate the up- and downstream models separately, as described above.

The brain volume regression model also demonstrates that both the uncertainty in the image and the propagated uncertainty in the final prediction are positively correlated with the acceleration factor (see Figure \ref{fig:scatterboth} right and Figure \ref{brain_overview}). Higher acceleration rates lead to increasing uncertainty in the data, which returns in a higher variance of the prediction and lower structural similarity compared to the ground truth image (see Table \ref{brainmc256}). This increased uncertainty is propagated by the downstream regression model. Higher upstream uncertainty yields a higher propagated variance $\sigma^2_{\hat{y}}$ and a higher joint variance $\var[y]$.\linebreak Both of them are given as the square root of the average over the dataset \linebreak($\sqrt{\sigma^2_{\hat{y}}}$ and  $\sqrt{\var[y]}$). Beyond the uncertainty estimations, the sparser input data also leads to reduced model performance, as the average L$_1$ (Manhattan) and L$_2$ (Euclidean) distances between the prediction and the ground truth brain volumes rise with higher undersampling factors. We show the results for the patient sex classification pipeline in the supplementary material.
\section{Conclusions}
To the best of our knowledge, this is the first work to demonstrate how uncertainty can be propagated through medical imaging pipelines consisting of cascades of deep learning models. Our method quantifies the models' individual contributions to the joint uncertainty and be consequently aggregates them to a joint uncertainty measure.
In extensive experiments, we have shown that our method can be integrated into real-world clinical image processing pipelines. 

At the moment, our method does not capture the spatial correlation between the uncertainty of pixels. Future work could extend our method to handle probability distributions beyond pixel-wise independent normal distributions. Additionally, it would be valuable to incorporate epistemic uncertainty into the method. One major challenge that remains unresolved is the calibration of the pipeline's uncertainty. Moreover, to ensure meaningful comparisons between uncertainty propagation techniques and effectively evaluate different pipelines, the establishment of a well-defined metric is imperative.

Ultimately, we envision that our method will allow clinicians to assess and apportion all sources of aleatoric uncertainty within the medical imaging pipeline, increasing their confidence when deploying deep learning in clinical practice.
%
%
\subsection*{Acknowledgments}This research has been funded by the German Federal Ministry of Education and Research under project „NUM 2.0“ (FKZ: 01KX2121). Data collection and sharing for this project was funded by the Alzheimer's Disease Neuroimaging Initiative (ADNI) (National Institutes of Health Grant U01 AG024904) and DOD ADNI (Department of Defense award number W81XWH-12-2-0012)
\bibliographystyle{splncs04}
\bibliography{mybibliography}
\end{document}


\title{Propagation and Attribution of Uncertainty in Medical Imaging Pipelines\\ Supplementary Material}
\authorrunning{L. F. Feiner et al.}
\author{
    Leonhard F. Feiner
    \and Martin J. Menten
    \and Kerstin Hammernik
    \and Paul Hager
    \and Wenqi Huang
    \and Daniel Rueckert
    \and Rickmer F. Braren
    \and Georgios Kaissis
}
\institute{
}
\maketitle
\section{Reconstruction of Brain MR Images and Classification of Sex}
\label{appendix:brainsex}
\begin{figure}[ht!]
\centering
\input{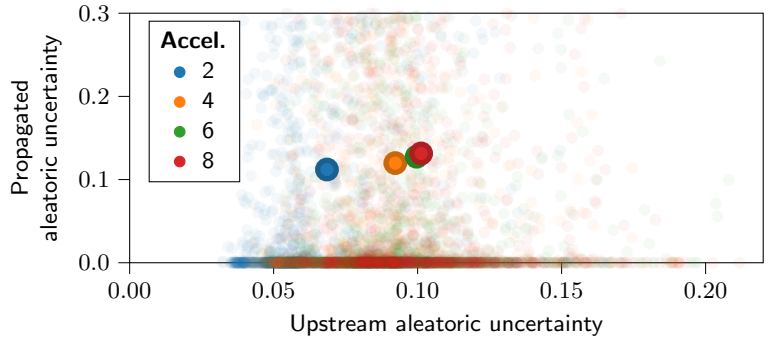}
    \caption{Higher acceleration factors (color) lead to higher reconstruction uncertainty (x-axis) and, through propagation via our method, contribute to a higher uncertainty of the final pipeline prediction of the patient's sex. The propagated portion of the output uncertainty is given as mutual information (y-axis). Large dots are aggregate values.
    }
 \label{fig:scatterbrainsex}
\end{figure}
\begin{table}
\centering
\caption{Quantitative measures characterizing the input k-space data, reconstructed images from the upstream model and final output from the downstream model that predicts the patient's sex.}
\label{fig:brainsexmc256}
\begin{tabular}{c c|c c|c c c c}
\multicolumn{2}{c|}{k-space} & \multicolumn{2}{c|}{Reconstructed image} & \multicolumn{4}{c}{Classification output} \\
Accel.& C. Frac.&\hphantom{--}SSIM\hphantom{--}& $\sqrt{\var[\textbf{x}]}$&\hphantom{-}ACC&\hphantom{-}$\mutual \left[y,\textbf{x}\right|\textbf{z}]$ & \hphantom{-}
$\expectation\left[\entropy \left[{y}|\textbf{x},\textbf{z}\right]\right]$
\hphantom{-}
&$\entropy \left[\expectation\left[{y}|\textbf{z}\right]\right]$ \\
\hline
2 &0.16 & 0.714 & 0.069 & 0.880 & 0.052 & 0.064 & 0.116 \\
4 &0.08 & 0.567 & 0.092 & 0.874 & 0.057 & 0.057 & 0.114 \\
6 &0.06 & 0.502 & 0.100 & 0.873 & 0.063 & 0.059 & 0.122 \\
8 &0.04 & 0.450 & 0.101 & 0.845 & 0.067 & 0.063 & 0.130 \\
\end{tabular}
\end{table}
\section{Supplement to Classification Model}
\label{appendix:classifier}
\leavevmode
\begin{figure}[ht!]
\centering
\input{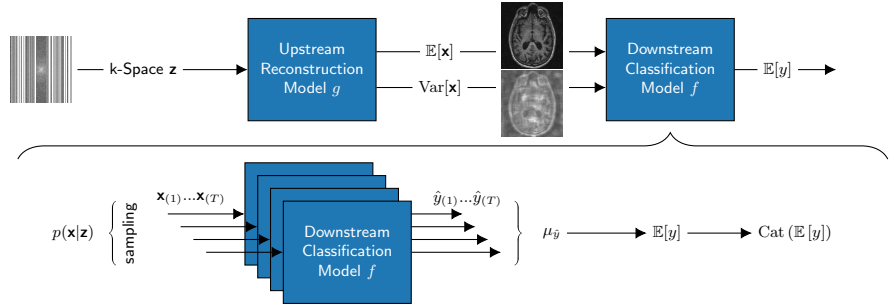}\\
\begin{tikzpicture}[scale=0.68, every node/.style={scale=0.68}]
{
    \node (downstreamm1)   [model]   { Downstream\\ Classification\\ Model $f$};
    \draw[->]  (downstreamm1.east) -- ++(1.75, 0);
    \node (downstreamm2)   [model,  below of=downstreamm1, xshift=0.25cm, yshift=0.75cm]   { Downstream\\ Classification\\ Model $f$};
    \draw[->]  (downstreamm2.east) -- ++(1.75, 0);
    \node (downstreamm3)   [model,  below of=downstreamm2, xshift=0.25cm, yshift=0.75cm]   { Downstream\\ Classification\\ Model $f$};
    \draw[->]  (downstreamm3.east) -- ++(1.75, 0);
    \node (downstreamm4)   [model,  below of=downstreamm3, xshift=0.25cm, yshift=0.75cm]   { Downstream\\ Classification\\ Model $f$};
    \draw[->]  (downstreamm4.east) -- ++(1.75, 0);
    \draw[->] ([xshift=-1.5cm]downstreamm1.west) -- (downstreamm1);
    \draw[->] ([xshift=-1.5cm]downstreamm2.west) -- (downstreamm2);
    \draw[->] ([xshift=-1.5cm]downstreamm3.west) -- (downstreamm3);
    \draw[->] ([xshift=-1.5cm]downstreamm4.west) -- (downstreamm4);

    \draw [decorate,decoration = {calligraphic brace,raise=10pt}] ([xshift=2.25cm,yshift=0.25cm]downstreamm1.east) -- ++(0, -1.25);
    \draw [decorate,decoration = {calligraphic brace,raise=10pt, mirror}] ([xshift=-2cm,yshift=0.25cm]downstreamm1.west) -- ++(0, -1.25);
    \node (input_distributution) [data, left of=downstreamm1, xshift=-3.6cm, yshift=-0.375cm] {$p(\textbf{x}|\textbf{z})$};
    \node (sampling) [data, left of=downstreamm1, xshift=-2.5cm, yshift=-0.375cm, rotate=90] {sampling};

    \node (input_sample) [txt, left of=downstreamm1, xshift=-1.3cm, yshift=0.3cm] {$\textbf{x}_{(1)} ... \textbf{x}_{(T)}$};
    \node (input_sample) [txt, right of=downstreamm1, xshift=2.1cm, yshift=0.3cm] {$\hat{y}_{(1)} ... \hat{y}_{(T)}$};

    \node (output_mean) [data, right of=downstreamm1, xshift=3.75cm, yshift=-0.375cm] {$\mu_{\hat{y}}$};
    \node (output_expectation) [data, right of=output_mean, xshift=1.25cm] {$\expectation[y]$};
    \draw[->] (output_mean) -- (output_expectation);

    \node (output_distribution) [data, right of=output_expectation, xshift=1.5cm] {$\categorical\left(\expectation\left[y\right]\right)$};
    \draw[->] (output_expectation) -- (output_distribution);
}
\end{tikzpicture}
\caption{An example of an imaging pipeline consisting of an upstream MR image reconstruction model $g$ and a downstream classification model  $f$. Both models predict a measure of heteroscedastic uncertainty. Our method allows the propagation of the mean and variance outputs of the upstream model through the downstream classification model. Finally, it combines the upstream uncertainty with the downstream uncertainty and yields a categorical distribution.} \label{fig:downstream_classification}
\end{figure}
\noindent\textbf{Monte Carlo Approximations of Uncertainty Metrics:}
The entropy of the predicted confidence is used as uncertainty metric for classification tasks. It is calculated as follows:
\begin{align*}
\entropy \left[y|\textbf{z}\right]
&=-\sum^C_{c=1}\expectation\left[p\left(y=c|\textbf{x}\right)\right] \log\expectation\left[p\left(y=c|\textbf{x}\right)\right]\\
&\approx-\sum^C_{c=1}\left(\frac{1}{T}\sum^T_{t=1}p\left(y=c|\textbf{x}_{(t)}\right)\right) \log\left(\frac{1}{T}\sum^T_{t=1}p\left(y=c|\textbf{x}_{(t)}\right)\right).
\end{align*}
The mutual information of the propagated uncertainty and the uncertainty of the downstream model denotes the contribution of the propagated uncertainty to the joint uncertainty. It is expressed as follows:
\begin{align*}
\mutual \left[y,\textbf{x}\right|\textbf{z}]
&= \entropy \left[y|\textbf{z}\right]-\entropy  \left[y|\textbf{x}, \textbf{z}\right]\\
&\approx-\sum^C_{c=1}\left(\frac{1}{T}\sum^T_{t=1}p\left(y=c|\textbf{x}_{(t)}\right)\right) \log\left(\frac{1}{T}\sum^T_{t=1}p\left(y=c|\textbf{x}_{(t)}\right)\right)\\
&\hspace{11pt}+\frac{1}{T}\sum^T_{t=1}\sum^C_{c=1}p\left(y=c|\textbf{x}_{(t)}\right) \log \left(p\left(y=c|\textbf{x}_{(t)}\right)\right).
\end{align*}